%% file: paper.tex
\documentclass[letterpaper, 10 pt, conference]{ieeeconf}
\IEEEoverridecommandlockouts    
\input{stachnisslab-latex}


\usepackage{threeparttable}
\usepackage{multirow} 
\usepackage{url}

\usepackage[capitalize]{cleveref}
\crefname{section}{Sec.}{Secs.}
\Crefname{section}{Section}{Sections}
\Crefname{table}{Table}{Tables}
\crefname{table}{Tab.}{Tabs.}

\title{\LARGE \bf TSCM: A Teacher-Student Model for Visual Place Recognition \\
Using Cross-Metric Knowledge Distillation }

\author{Yehui Shen \and Mingmin Liu  \and Huimin Lu \and Xieyuanli Chen 
  \thanks{Y. Shen is with Faculty of Robot Science and Engineering, Northeastern University, China. X. Chen and H. Lu are with the College of Intelligence Science and Technology, National University of Defense Technology, China. M. Liu is with SIASUN Robot\& Automation Co., Ltd, China. }%
  \thanks{Corresponding author: Xieyuanli Chen (xieyuanli.chen@nudt.edu.cn)} 
  \thanks{This work has partially been funded by the National Science Foundation of China under Grant U1913202, U22A2059, U20A20197, and 62203460, Fund for key Laboratory of Space Flight Dynamics Technology (Num 2022-JYAPAF-F1028), and Young Elite Scientists Sponsorship Program by CAST (No. 2023QNRC001).
  }%
}

\begin{document}
\maketitle
\thispagestyle{empty}
\pagestyle{empty}

\begin{abstract}
Visual place recognition (VPR) plays a pivotal role in autonomous exploration and navigation of mobile robots within complex outdoor environments. While cost-effective and easily deployed, camera sensors are sensitive to lighting and weather changes, and even slight image alterations can greatly affect VPR efficiency and precision. Existing methods overcome this by exploiting powerful yet large networks, leading to significant consumption of computational resources. In this paper, we propose a high-performance teacher and lightweight student distillation framework called TSCM. It exploits our devised cross-metric knowledge distillation to narrow the performance gap between the teacher and student models, maintaining superior performance while enabling minimal computational load during deployment. We conduct comprehensive evaluations on large-scale datasets, namely Pittsburgh30k and Pittsburgh250k. Experimental results demonstrate the superiority of our method over baseline models in terms of recognition accuracy and model parameter efficiency. Moreover, our ablation studies show that the proposed knowledge distillation technique surpasses other counterparts. The code of our method has been released at \url{https://github.com/nubot-nudt/TSCM}.
\end{abstract}

\section{Introduction}
\label{sec:intro}

Place recognition plays a significant role in the autonomous navigation of mobile robots, allowing them to locate the global place within previously mapped environments~\cite{yin2023arxiv, yin2022arxiv}. Existing approaches can be categorized into visual place recognition (VPR) and LiDAR place recognition (LPR). Though LPR methods~\cite{chen2020rss, ma2022ral, ma2022tie} have demonstrated good performance in outdoor environments, 3D LiDAR sensors are comparatively more costly and power-intensive. In contrast, cameras offer the advantages of cost-efficiency and abundant perceptual data, leading to growing research interest in outdoor place recognition~\cite{zhang2021pr, Masone2021access}.
 
Current VPR methods~\cite{Hausler2021cvpr} typically compress the query image into a global descriptor that is then used to determine the current location by comparing it with stored data in the database. However, visual sensors are sensitive to environmental changes, such as variations in lighting, weather conditions, and occlusions, which can negatively impact the robustness of these descriptors, resulting in poor VPR performance. Therefore, the main focus in VPR research is to develop a robust network capable of generating stable descriptors. Unfortunately, such networks tend to be complex, with a high number of parameters, and they often operate at computational speeds lower than the sensor's frame rate, which limits their practical applicability~\cite{yu2019tnnls}.

To ﬁll this gap, Hui~\etalcite{hui2022tip} employed knowledge distillation to enable knowledge transfer from the powerful teacher network to the lightweight student network. However, the performance of the student network fell significantly short of that of the teacher network. 
Cai~\etalcite{cai2022iros} introduced a self-learning approach to improving the student network in acquiring the teacher network's knowledge. While this approach enhanced the student network's performance, it did not reduce the size of the student network, making is unsuitable for mobile robots with limited onboard resources.

\begin{figure}[t]
  \centering
  \includegraphics[width=\linewidth]{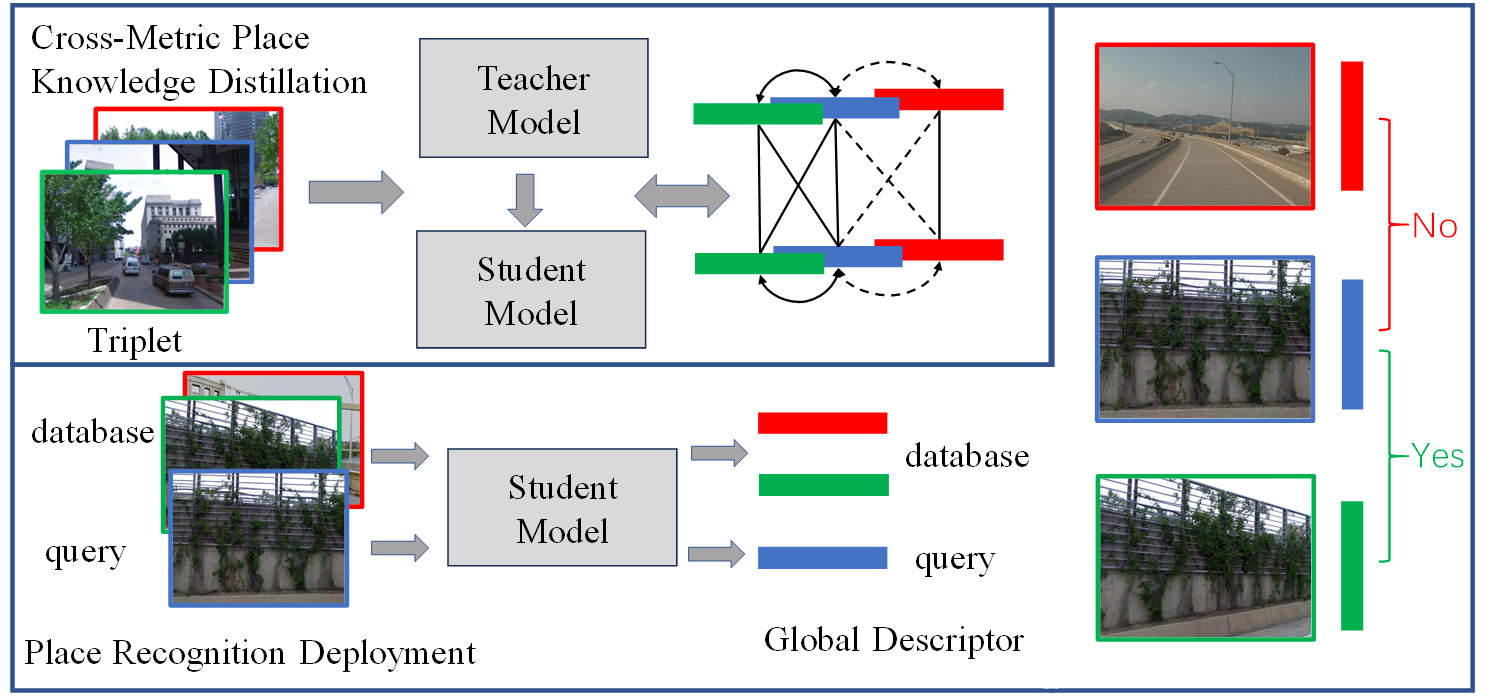}
  \caption{Our model consists of cross-metric knowledge distillation and place recognition deployment. It uses cross-metric learning to transfer knowledge from teacher to student offline. During online inference, it uses the lightweight student model to generate descriptors from the input image and identify potential places by comparing them against those stored in the database.}
  \label{fig:model}
\end{figure}

In this paper, we introduce a novel concept called cross-metric knowledge distillation (KD) and apply it to VPR within our teacher-student network. Our network comprises a teacher and a student model. The teacher model integrates components from ResNet~\cite{he2016cvpr}, vision transformer~\cite{dosovitskiy2021arxiv}, and inter-transformer~\cite{ma2023tii}, resulting in increased complexity and superior performance compared to state-of-the-art baselines in VPR. In contrast, the student network is designed to be lightweight, retaining only essential components. We transfer place knowledge from the powerful teacher to the lightweight student through our proposed cross-metric distillation. This approach empowers the lightweight student network to match or exceed the teacher's performance while reducing inference time, enabling fast and accurate VPR in real-world applications, as illustrated in~\cref{fig:model}.
We evaluated our model on two large outdoor datasets, Pittsburgh30k and Pittsburgh250k. The experimental results demonstrate that our model outperforms baseline models in terms of both recognition accuracy and parameter efficiency. Furthermore, we provide multiple ablation studies showing that our proposed distillation method surpasses other distillation techniques.

In summary, the contributions of this paper are threefold:

(i) We introduce TSCM, a novel cross-metric knowledge distillation approach applied to VPR, which enables the student model to even outperform the teacher model.

(ii) Both our devised new teacher and student networks achieve better performance in terms of VPR accuracy than the state-of-the-art baseline methods;

(iii) While achieving superior performance, our method is more lightweight than the baseline, displaying outstanding computational efficiency. Our student model generates one descriptor and finds a matching within a 10\,k-image database in less than 2\,ms, achieving real-time performance.

\section{Related Work}
\label{sec:related}

Localization is a crucial capability for robots, commonly achieved through Simultaneous Localization and Mapping (SLAM). However, over time, inevitable drift adversely impacts localization accuracy. Responding to this challenge, researchers have increasingly focused on place recognition. Place recognition aids in loop-closure detection and re-localization by identifying previously visited locations and mitigating localization drift~\cite{chen2020rss, chen2021auro, wang2020iros, siam2017icra}. This paper primarily centers on visual place recognition (VPR).

Early work by Milford~\etalcite{Milford2012icra} introduced SeqSLAM for outdoor VPR. They employed handcrafted features to calculate the most suitable candidate among sequences.  Vysotska and Stachniss~\cite{vysotska2019ral} exploit multi-sequence maps as well as sequential image information to achieve effective and robust VPR.
With the development of neural network technology, there has been a surge in the development of learning-based methods. Among these, NetVLAD~\cite{arandjelović2016cvpr}, a Convolutional Neural Network (CNN)-based architecture, stands out as a pivotal advancement. It facilitates end-to-end training for place recognition, thereby achieving considerable success. Subsequently, many NetVLAD-based approaches have emerged~\cite{yu2020tnnls, Khaliq2020tro, Hausler2021cvpr,li2023ral,Khaliq2022ral}. 
 
With the advent of the Transformer~\cite{vaswani2017nips}, an increasing number of Transformer-based methods have come to the forefront. Wang~\etalcite{wang2023tcsvt} introduced a hybrid CNN-Transformer feature extraction network, addressing the inherent locality limitation of Convolutional Neural Networks (CNNs) by incorporating the Vision Transformer to model image contextual information and dynamically combining task-related features. Hou~\etalcite{hou2022icra} used Transformer-based networks for feature extraction, presenting a hierarchical transformer for VPR. 

Knowledge distillation was initially introduced for image classification by Hinton~\etalcite{hinton2015arxiv}. This technique utilizes a powerful, well-trained teacher model to guide the training process of a lighter-weight student model. 
The goal is to simplify the model while preserving the original teacher model's accuracy. Due to the success of knowledge distillation, many related works have emerged.~\cite{cho2019iccv, park2019cvpr-rkd, zhao2022cvpr}. Oki~\etalcite{oki2020ijcnn} introduced the concept of metric learning into knowledge distillation, leveraging pairs or triplets of training samples to bring the student model closer to the teacher model, which is closely aligned with the concept of cross-metric knowledge distillation proposed in this paper. In recent years, researchers have also applied knowledge distillation to place recognition. Shen~\etalcite{Shen2023cvpr} proposed StructVPR,  a novel training architecture for VPR, aiming to enhance structural knowledge within RGB global features, thereby enhancing feature stability in dynamic environments. Cui~\etalcite{Cui2023ral} introduced a continual contrastive learning method named CCL to address the issue of catastrophic forgetting and enhance the overall robustness of LPR approaches. This method employs knowledge distillation to update the model, enabling it to sustain effective place recognition over extended periods and across diverse environments. Cai~\etalcite{cai2022iros} introduced STUN, a self-teaching framework that predicts place and estimates prediction uncertainty simultaneously using knowledge distillation without reducing the size.

In contrast to the abovementioned approaches, we introduce a novel knowledge distillation technique called cross-scale knowledge distillation, which helps us achieve a good balance between performance and model simplicity for VPR.

\section{Our Approach}
\label{sec:main}

In this section, we present the detailed network architecture of our teacher-student models, TSCM, and how we apply the devised cross-metric KD to VPR.

\begin{figure*}[t]
  \centering
  \includegraphics[width=0.85\linewidth]{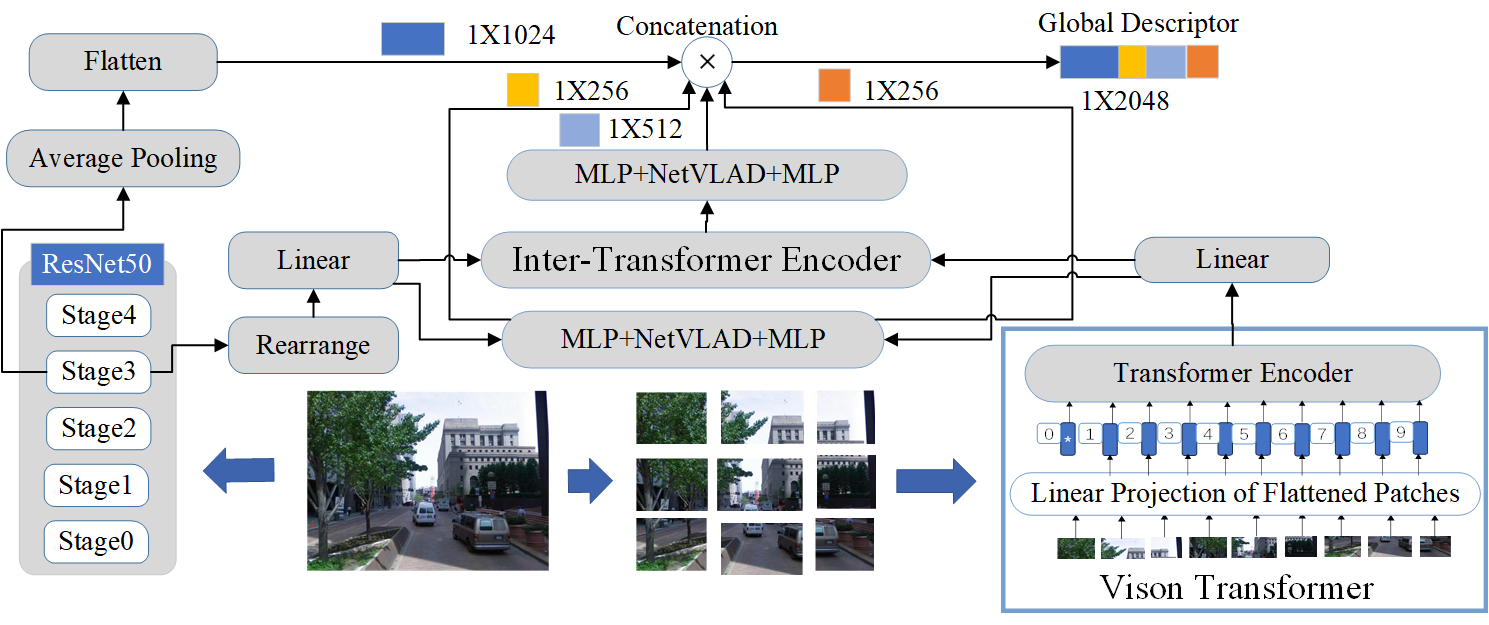}
  \caption{The pipeline overview of our proposed teacher network. It processes input images through a multi-stage feature extraction process using ResNet and Vision Transformer (ViT). The extracted features are further processed by NetVLAD and MLPs to create initial descriptors. The final global descriptor is the concatenation of features from all branches.}
  \label{fig:Teacher Network}
\end{figure*}

\subsection{Teacher-Student Model}
The architecture of our designed teacher network is presented in~\cref{fig:Teacher Network}. It fuses powerful components from ResNet~\cite{he2016cvpr}, vision transformer~\cite{dosovitskiy2021arxiv}, inter-transformer~\cite{ma2023tii} to increase complexity, enabling better learning of place knowledge in complex outdoor environments and generating distinctive place descriptors.
It first exploits ResNet~\cite{he2016cvpr} to extract high-level and fine-grained features from the input image.  While deeper network architectures have the potential to extract more intricate features, they often encounter training complexities. ResNet effectively addresses these challenges through the use of residual connections. Therefore, we employ several ResNet blocks as the backbone network, as illustrated in~\cref{fig:Teacher Network}. We exclude its final stage to enable the teacher model to allocate more general features relevant to places, which is verified to be more beneficial than the features of the last layer in our ablation studies.
Previous research~\cite{cai2022iros} did not utilize the middle features directly for the final descriptor, but our findings show that using these features improves outcomes. As a result, we preserve the $(1\times1024)$ features from the ResNet branch and incorporate them into the final global descriptor aggregation process. This insight is also verified in~\cref{fig:aba}.

\begin{figure}[t]
  \centering
  \includegraphics[width=\linewidth]{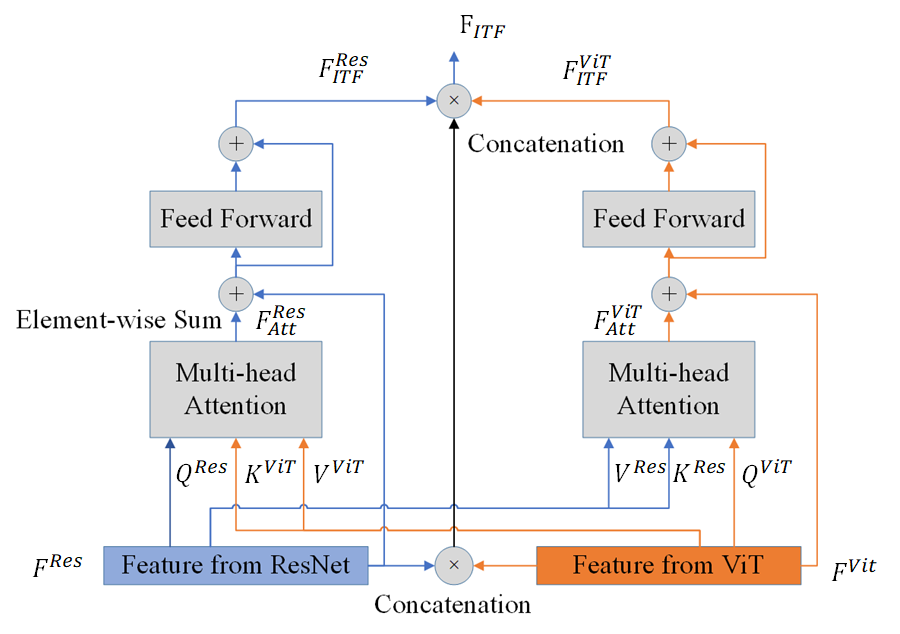}
  \caption{The structure of the Inter-Transformer Encoder}
  \label{fig:InterTF}
\end{figure}

While ResNet has demonstrated good performance, its feature extraction tends to overlook interrelationships among distinct regions within an image. In contrast, Vision Transformer (ViT)~\cite{dosovitskiy2021arxiv} extracts both global and local features from the image while simultaneously capturing correlation information among various image regions. It establishes connections among these disparate regions, enhancing its capacity to capture global and contextual information. Therefore, we also employ ViT to generate global features from the input image. After acquiring fine-grained features of $(1\times256)$ from ResNet and global features of the same size from ViT, we combine them to create more representative image features, generating distinctive global descriptors.

To this end, we introduce an inter-transformer encoder to fuse the features extracted by ResNet and ViT as shown in~\cref{fig:InterTF}. It has two branches, each with a multi-head attention(MHSA)  and a feed-forward network(FFN). The outputs of each branch are finally fused by a NetVLAD and MLP module to obtain the initial descriptor of $(1\times512)$.
The MHSA of Inter-Transformer in~\cref{fig:InterTF} can be mathematically formulated as follows: 
\begin{equation} 
\begin{split}
F^{\text{Res}}_{\text{Att}} & = \text{Attention}(Q^{\text{Res}},K^{\text{ViT}},V^{\text{ViT}}) \\
             & = \text{softmax}\left(\frac{Q^{\text{Res}}K^{\text{ViT}}}{\sqrt{d_k}}\right)V^{\text{ViT}}  
\end{split}
\end{equation}
where $Q^{\text{Res}}$ is substantially the query split of the feature from ResNet. $K^{\text{ViT}}$,$V^{\text{ViT}}$are the key and value splits of the feature from ViT $F^{\text{ViT}}$. $\sqrt{d_k}$ represents the dimension of splits. $F^{\text{Res}}_{\text{Att}}$ denotes the output feature from the MHSA in the ResNet branch. This feature is subsequently fed into the FFN to generate the attention-enhanced feature, facilitated by the following mathematical formulation:
\begin{equation} 
\begin{split}
F^{\text{Res}}_{\text{ITF}} = \text{FFN}(F^{\text{Res}}_{\text{Att}}+F^{\text{Res}})+F^{\text{Res}}_{\text{Att}}+F^{\text{Res}}   
\end{split}
\end{equation}

Likewise, we perform the same operations on the ViT branch.:
\begin{equation} 
\begin{split}
F^{\text{ViT}}_{\text{Att}} & = \text{Attention}(Q^{\text{ViT}},K^{\text{Res}},V^{\text{Res}}) \\
             & = \text{softmax}\left(\frac{Q^{\text{ViT}}K^{\text{Res}}}{\sqrt{d_k}}\right)V^{\text{Res}}  
\end{split}
\end{equation}
\begin{equation} 
\begin{split}
F^{\text{ViT}}_{\text{ITF}} = \text{FFN}(F^{\text{ViT}}_{\text{Att}}+F^{\text{ViT}})+F^{\text{ViT}}_{\text{Att}}+F^{\text{ViT}}   
\end{split}
\end{equation}

Features $F^{\text{Res}}_{\text{ITF}}$ and $F^{\text{ViT}}_{\text{ITF}}$ are concatenated to form the composite feature $F_{\text{ITF}}$. Subsequently, this feature $F_{\text{Inter}}$ is fed into the NetVLAD-MLPs combos~\cite{arandjelović2016cvpr} to generate a descriptor. The final global descriptor is created by integrating this descriptor with those from other branches. While the teacher network gathers maximum information using multiple complex components, including potentially redundant knowledge, this does not affect deployment, as a lighter student network is deployed into practical applications.

\begin{figure}[t]
  \centering
  \includegraphics[width=\linewidth]{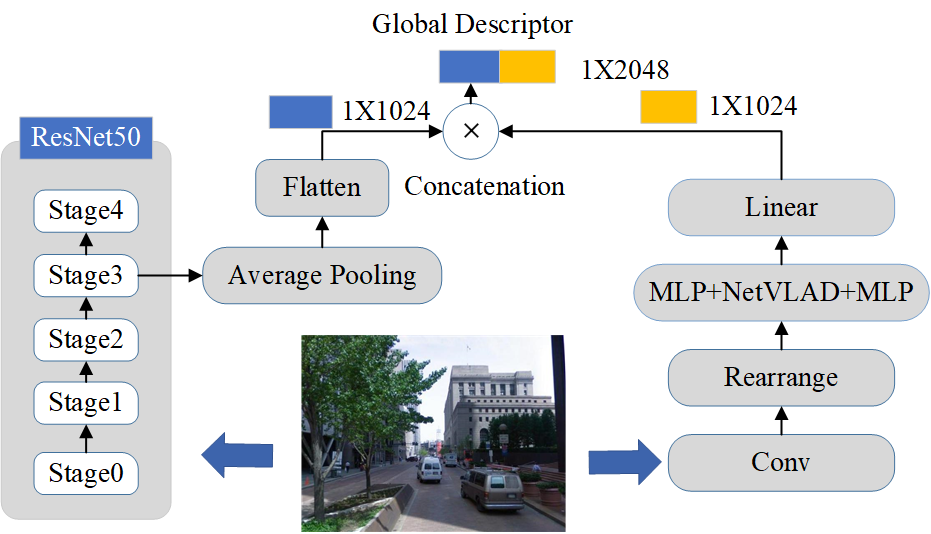}
  \caption{The overview of the student network}
  \label{fig:Student}
\end{figure}

The overview of our student network is depicted in~\cref{fig:Student}. Unlike the complex architecture of the teacher network, the student network has a much simpler structure, as illustrated in~\cref{fig:aba}. The ResNet branch is pivotal in place recognition, so we keep it in the student network. To simplify the student network, we replace both the ViT and the Inter-Transformer with a layer of convolutional layers. Our student network is relatively smaller but still includes multiple branches, allowing it to learn various types of place knowledge from the teacher network through cross-metric knowledge distillation, explained in the next section.

\subsection{Cross-Metric Knowledge Distillation}

\begin{figure}[t]
  \centering
  \includegraphics[width=\linewidth]{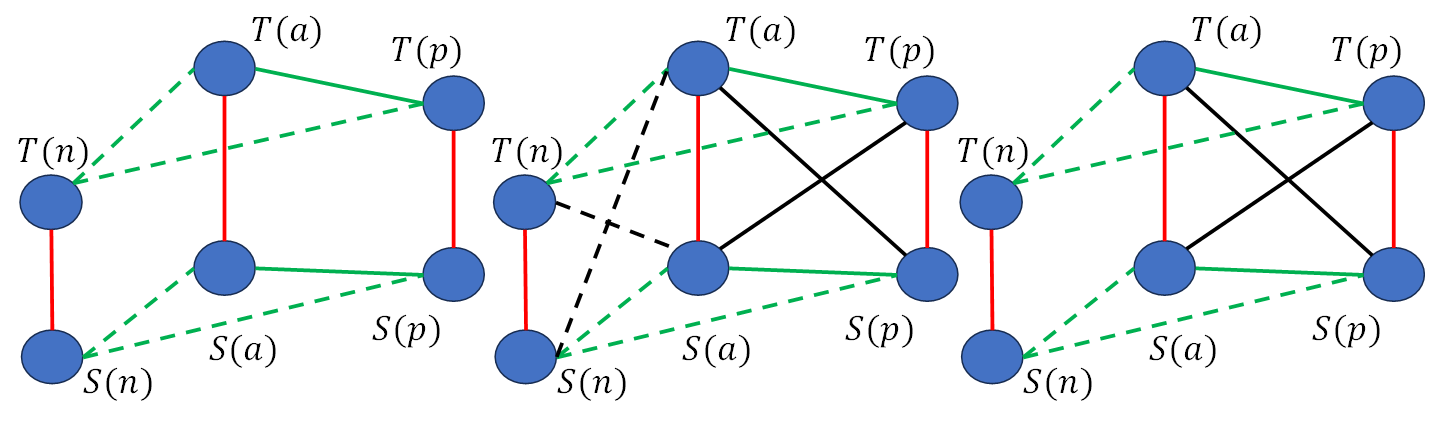}
  \caption{Comparison of different knowledge distillation (KD) strategies on triplet loss. $T(\cdot)$ represents the output of the teacher model, and $S(\cdot)$ represents the output of the student model. $a$, $p$, $n$ represent anchor, positive, and negative samples. Solid lines indicate distances to be reduced, while dashed lines indicate distances to be increased. The left figure shows the traditional KD method. The middle one is a fully connected knowledge distillation. The right figure illustrates our proposed strategy, which exploits constraints within each model and facilitates cross-model interactions, enhancing supervision for knowledge distillation in place recognition.}
  \label{fig:CMKD}
\end{figure}

Knowledge distillation (KD) transfers complex knowledge from a larger teacher model to a smaller student model while keeping the student compact. 
Existing KD methods for VPR typically use triplet loss~\cite{hoffer2015sbpr}, as shown in the left figure of~\cref{fig:CMKD}. $T(\cdot)$ represents the teacher model, and $S(\cdot)$ represents the student model. $a$, $p$, and $n$ correspond to the anchor, positive, and negative samples. Solid lines indicate distances to be reduced, while dashed lines indicate distances to be increased.
Triplet loss is well-suited for supervising VPR tasks, as it helps find similar places and distinguish different places.
During KD, the teacher model remains fixed, and the triplet loss is applied to the student model as follows:
\begin{equation} 
\begin{split}
L_{\text{Tri}} = \max\left\{ d(S(a) - S(p))-d(S(a) - S(n))+m),0 \right\},
\end{split}
\label{con:explanation}
\end{equation}
where $m$ is the margin.

Traditional KD aligns outputs of the student model with those of the teacher, keeping distances between $T(a)$, $T(n)$, and $T(p)$ unchanged, known as soft targets (red lines in~\cref{fig:CMKD}), and the corresponding loss function is:
\begin{equation} 
\begin{split}
L_{\text{soft}} & = \sum^{\substack{N}}_{\substack{i}} d(S(a_i) - T(a_i))+d(S(p_i) - T(p_i)) \\
         &+d(S(n_i) - T(n_i)).
\end{split}
\label{con:soft}
\end{equation}

In VPR, our objective is to ensure that positive sample outputs are close to the anchor while negative sample outputs are far from it, referred to as hard targets (illustrated by the green lines in~\cref{fig:CMKD}). This goal should be consistent whether dealing with teacher samples, student samples, or interactions between them.
Traditionally, the triplet loss focuses only on the relationship between the outputs of the student model, overlooking the relationship between the outputs of the student and teacher models (e.g., the relationship between $S(a)$ and $T(p))$. 
One can apply the triplet loss within a single model to achieve the desired hard targets ($L_{\text{Tri}} = L_{\text{hard}}$). However, there is limited research on achieving these hard targets between the teacher and student models.

To fill in this gap, we investigated the relationships between samples in both the teacher and student models during the distillation process. We aimed to bring a positive student sample, S(p), closer to the teacher anchor, T(a), while pushing negative student samples, S(n), away from T(a), and vice versa. To capture these relationships, we introduced black lines in addition to the red and green lines, as depicted in the second figure of~\cref{fig:CMKD}, to represent the distances between samples across teacher and student. This can be formulated as:
\begin{equation}  
\begin{split}
L_{\text{cm-full}} & = \max \left\{ d(S(a) - T(p)) + d(S(p) - T(a)) \right. \\
       & \quad \left.- d(S(a) - T(n)) - d(S(n) - T(a)) + m, 0 \right\}.
\end{split}
\label{eq:loss}
\end{equation}

Ideally, using~\cref{eq:loss} as the loss for VPR would maximize information utilization during KD. However, including all sample distances presents a challenge: We aim to separate $S(a)$ from $T(n)$ and $S(n)$ from $T(a),$ but this may inadvertently cause $S(a)$ to diverge from $T(a)$ and $S(n)$ to diverge from $T(n),$ contradicting our soft targets (red solid line). To address this, we propose our cross-metric KD, which discards the dashed black lines and maintains the solid black lines as cross-model constraints, formulated as:
\begin{equation} 
\begin{split}
L_{\text{cm}} =\sum^{\substack{N}}_{\substack{i}} d(S(a_i) - T(p_i))+d(S(p_i) - T(a_i))
\end{split}
\label{con:cm}
\end{equation}
As $S(a)$ approaches $T(p)$ and $S(p)$ approaches $T(a)$, the distances between $S(a)$ and $T(a)$, $S(p)$ and $T(p)$, and $S(a)$ and $S(p)$ tend to decrease. As illustrated in the right figure of~\cref{fig:CMKD}, our design benefits both soft and hard targets, thus increasing the KD performance specifically in VPR tasks.

Taking into account the aforementioned considerations, we define the final loss function for our cross-metric KD for VPR as follows:
\begin{equation} 
\begin{split}
L_{\text{total}}  =  L_{\text{hard}}+L_{\text{soft}}+L_{\text{cm}}.
\end{split}
\label{con:KD}
\end{equation}

\section{Experimental Evaluation}
\label{sec:exp}
The main focus of this work is a teacher-student network exploiting our devised cross-metric knowledge distillation (KD) for fast and accurate visual place recognition (VPR). 
We present our experiments to show the capabilities of our method and  support our key claims, which are:
(i) TSCM introduces the concept of cross-metric KD to VPR, allowing our smaller student model to perform similarly, and sometimes even better, than the larger teacher model;
(ii) TSCM attains superior recognition accuracy while maintaining a more lightweight model in comparison to the state-of-the-art baseline methods;
(iii) TSCM demonstrates exceptional computational efficiency, compressing images into descriptors in 1.3\,ms and finding a matching in under 0.6\,ms per query using a 10\,k-image database.

\subsection{Experimental Setup}

1) Dataset: Our evaluation of TSCM and other methods is conducted on the Pittsburgh dataset~\cite{Torii2015tpami}, a comprehensive image repository sourced from Google Street View. The original dataset encompasses 250\,k training samples and 24\,k test query samples. For the training procedure, we employ the same partitioning scheme as NetVLAD~\cite{arandjelović2016cvpr}, using a reduced subset of 10\,k samples for training and validation while employing Pittsburgh30k with 10\,k test data and Pittsburgh250k with 84\,k test data for testing. Notably, these images have been captured at various times of the day and span multiple years.

2) Configuration: In TSCM, the Vision Transformer (ViT) is pre-trained at a resolution of 224x224, with a patch size of 16. For the Inter-Transformer, our setup aligns with that of CVTNet~\cite{ma2023tii}, featuring the following specifications: an embedding dimension of $d_{\text{model}}$ = 512, several heads of $n_{\text{heads}}$= 4, and an intermediate dimension for the feed-forward layer of $d_{\text{ff}}$ = 1024. In the case of NetVLAD following the Inter-Transformer, we configure it with an input feature vector dimension of $\text{feature}\_{\text{size}}$ = 512, an output feature dimension of $d_{\text{output}}$ = 512, and a total of $d_K$ = 64 clusters.For other instances of NetVLAD, we adjust the input feature vector dimension to $\text{feature}\_{\text{size}}$  = 256, with an output feature dimension of $d_{\text{output}}$ = 256, while keeping the remaining parameters consistent with NetVLAD following the Inter-Transformer. During the network training phase, we adhere to the STUN setup~\cite{cai2022iros}, employing a batch size of 8, utilizing the Adam optimizer with an initial learning rate of $1\times10^{-5}$, which undergoes decay by a factor of 0.99 after each epoch, and applying a weight decay of 0.001. Our training infrastructure is hosted on a cloud server equipped with the following specifications: a single V100-SXM2-32GB (32GB) GPU, complemented by a 12-core Intel(R) Xeon(R) Platinum 8255C CPU @ 2.50GHz. The server environment operates on the PyTorch 1.10.0 image, utilizing Python 3.8 (Ubuntu 20.04) and CUDA 11.3.

3) Evaluation Metrics: Our experiments are designed to assess the recognition performance of our model and the impact of cross-metric knowledge distillation. We employ recall@N, mAP@N, and AP as metrics to measure recognition performance and recall@N as a metric to measure the impact of cross-metric knowledge distillation.
\begin{table}[t]
\footnotesize
\renewcommand\arraystretch{1.1}
\setlength{\tabcolsep}{2.5pt}
  \caption{Recognition performance on the Pittsburgh30k dataset}
  \centering
  \begin{tabular}{l|ccc}
    \toprule
     &r@1/5/10 $\uparrow$    &mAP@1/5/10 $\uparrow$   &AP $\uparrow$    \\
    \midrule
MC Dropout~\cite{taha2019arxiv} &0.302\,/\,0.523\,/\,0.611  &0.302\,/\,0.108\,/\,0.061  &0.463    \\
PFE w/o MLS~\cite{shi2019iccv} &0.550\,/\,0.805\,/\,0.876  &0.550\,/\,0.266\,/\,0.167  &0.690   \\
PFE w/ MLS~\cite{shi2019iccv} &0.444\,/\,0.680\,/\,0.764  &0.444\,/\,0.199\,/\,0.120  &0.655   \\
BTL~\cite{warburg2021iccv} &0.515\,/\,0.766\,/\,0.840  &0.515\,/\,0.252\,/\,0.158  &0.591    \\
STUN(teacher)~\cite{cai2022iros}  &0.548\,/\,0.802\,/\,0.877  &0.548\,/\,0.268\,/\,0.167  & 0.678  \\
STUN(student)~\cite{cai2022iros}  &0.613\,/\,0.840\,/\,0.898  &0.613 \,/\,0.280\,/\,0.171  &0.739   \\
TSCM(our teacher)   &\underline{0.730}\,/\,\underline{0.909}\,/\,\underline{0.942}  &\underline{0.730} / \textbf{0.334}\,/\,\textbf{0.204}  &\underline{0.862}  \\
TSCM(our student)    &\textbf{0.735}\,/\,\textbf{0.914}\,/\,\textbf{0.945} &\textbf{0.735}\,/\,\underline{0.329}\,/\,\underline{0.198}  &\textbf{0.864}  \\ 
   \bottomrule
  \end{tabular}
  \begin{tablenotes} 
\item \textbf{bold} denotes the best and \underline{underline} denotes the second best results.
\end{tablenotes}
  \label{tab:eva}
\end{table}
\begin{table}[t]
\footnotesize
\renewcommand\arraystretch{1.1}
\setlength{\tabcolsep}{2.5pt}
  \caption{Recognition performance on the Pittsburgh250k dataset}
  \centering
  \begin{tabular}{l|ccc}
    \toprule
     &r@1/5/10 $\uparrow$    &mAP@1/5/10 $\uparrow$   &AP $\uparrow$    \\
 \midrule
STUN(teacher)~\cite{cai2022iros}  &0.519\,/\,0.757\,/\,0.829  &0.519\,/\,0.235\,/\,0.143  &0.687\\
STUN(student)~\cite{cai2022iros}  &0.559\,/\,0.788\,/\,0.852  &0.559\,/\,0.249\,/\,0.150 &0.710   \\
TSCM(our teacher)   &\underline{0.680}\,/\,\underline{0.862}\,/\,\underline{0.900}  &\underline{0.680}\,/\,\textbf{0.299}\,/\,\textbf{0.177}  &\underline{0.846}  \\
TSCM(our student)    &\textbf{0.684}\,/\,\textbf{0.867}\,/\,\textbf{0.906} &\textbf{0.684}\,/\,\underline{0.294}\,/\,\underline{0.173}  &\textbf{0.847}  \\ 
 \bottomrule
  \end{tabular}
\begin{tablenotes} 
\item \textbf{bold} denotes the best and \underline{underline} denotes the second best results.
\end{tablenotes}
 \label{tab:eva_250k}
 \end{table}

\begin{figure}[t]
  \centering
  \includegraphics[width=\linewidth]{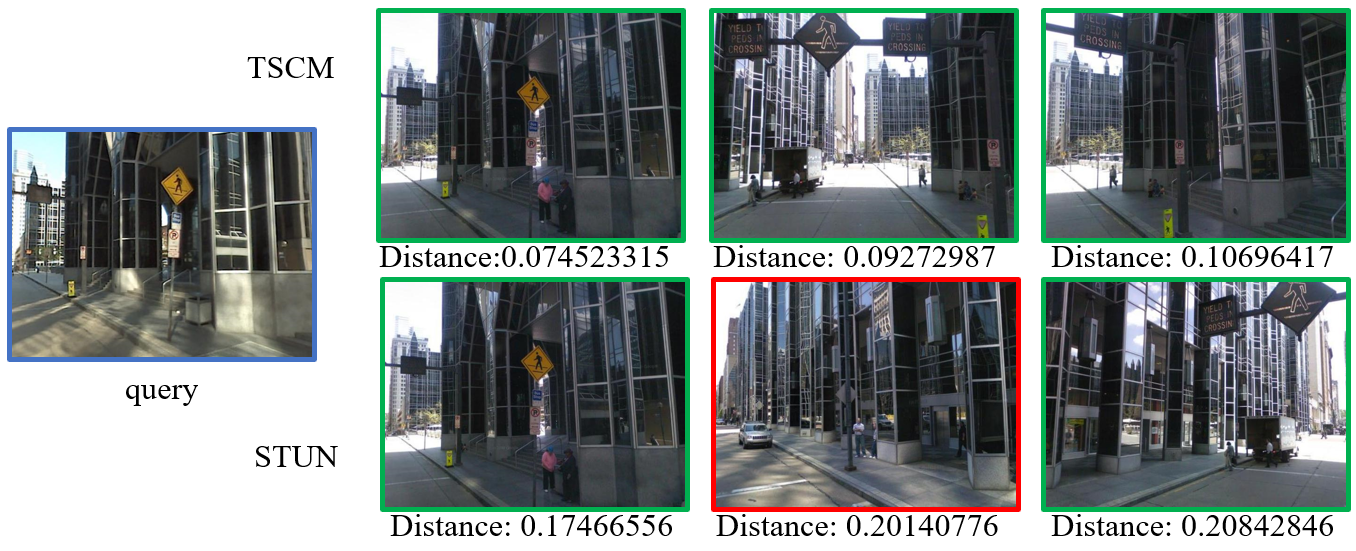}
  \includegraphics[width=\linewidth]{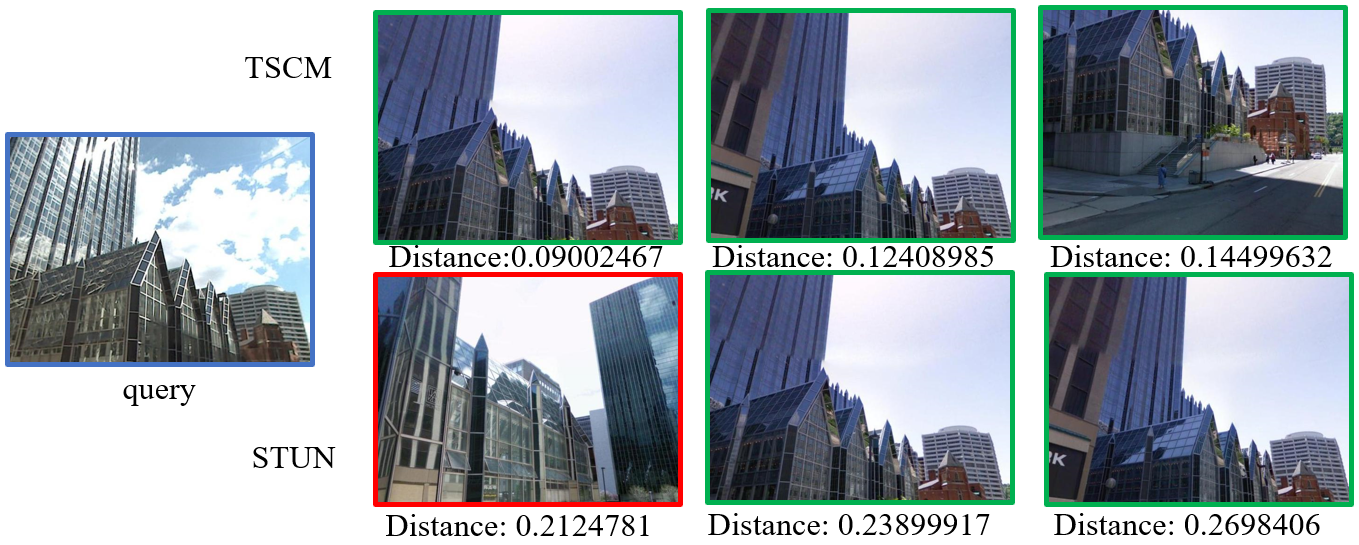}
  \vspace{-0.4cm}
  \caption{Comparing qualitative VPR results between ours and STUN}
  \label{fig:performance2}
  \vspace{-0.2cm}
\end{figure}

\subsection{VPR Performance}
The first experiment aims to assess the VPR performance of our approach. The experimental results support that TSCM achieves superior recognition accuracy compared to the baseline and our proposed KD strategy, allowing the smaller student model to perform similarly, and sometimes even better, than the larger teacher model. We follow the experimental setup the same as STUN~\cite{cai2022iros}.

\cref{tab:eva} presents the recognition performance of various methods on the Pittsburgh30k dataset. In terms of VPR performance, both our student and teacher method outperforms all other baseline methods. Notably, our student model outperforms our large teacher model in recall metrics while performing similarly in mAP and AP metrics. Our student model surpasses other baseline models in all metrics, including larger teacher models of baseline methods. 

To validate our method's generalization ability on a larger database, we compared it with STUN on Pittsburgh250k without retraining or fine-tuning. As depicted in~\cref{tab:eva_250k}, our method consistently achieves significantly better performance compared to the baseline methods. A more detailed comparison is also provided in the top-left figure of~\cref{fig:aba}. The recall and precision curves clearly illustrate that our method outperforms STUN on two datasets, affirming its superior performance.
We additionally provide more visualization results in~\cref{fig:performance2}. As can be seen, our TSCM consistently identifies true positives in VPR with smaller distances to the query compared to STUN, which produces incorrect matches.

\begin{table}[t]
\footnotesize
\renewcommand\arraystretch{1.1}
\setlength{\tabcolsep}{4.5pt}
\centering
\caption{Comparisons in params and runtime at database of 10\,k}
\begin{tabular}{l|ccccc}
\hline
 \multirow{2}{*}{ \centering model} &  Total  &  FLOPS & Descriptor  &  Matching & Total \\
& params & ($10^9$)  & generation & time & time \\ 
\hline
Our teacher & 194\,M  &81.626 &4.5\,ms&0.7\,ms &5.2\,ms\\
STUN's student &27\,M & 4.136 &1.3\,ms&0.6\,ms &1.9\,ms\\
Our student &13\,M &3.440  &1.3\,ms   &0.6\,ms &1.9\,ms\\
\hline
\end{tabular}
  \label{tab:para}
  \vspace{-0.2cm}
\end{table}

\begin{table}[t]
\footnotesize
\renewcommand\arraystretch{1.1}
\setlength{\tabcolsep}{2.2pt}
\centering
\caption{Effect of cross-metric knowledge distillation}
\begin{tabular}{cccccc|cc}
\hline
$L_{h}$&$L_{s}$ &$D_1$ &$D_2$&$D_3$ &$D_4$ & recall@1/5/10\,on\,30k & recall@1/5/10\,on\,250k\\
\hline
$\Delta$ &  &   &  &   &   & 0.730\,/\,0.909\,/\,0.942 &0.680\,/\,0.862\,/\,0.900       \\
\hline
$\surd$ & $\surd$ &   &   &   &   & 0.727\,/\,0.908\,/\,0.941 &0.679\,/\,0.863\,/\,0.898\\
\hline
$\surd$ & $\surd$&$\surd$&$\surd$&$\surd$&$\surd$&0.722\,/\,0.904\,/\,0.937  &0.671\,/\,0.851\,/\,0.890      \\
\hline
 $\surd$ & $\surd$&$\surd$&$\surd$&$\surd$&   &0.729\,/\,0.908\,/\,0.942 &0.680\,/\,0.864\,/\,0.890 \\
\hline
$\surd$ & $\surd$&$\surd$&$\surd$& &$\surd$& 0.725\,/\,0.911\,/\,0.943 & 0.676\,/\,0.859\,/\,0.896      \\
\hline
$\surd$ & $\surd$&$\surd$&$\surd$&   &  &\textbf{0.735}\,/\,\textbf{0.914}\,/\,\textbf{0.945} &\textbf{0.684}\,/\,\textbf{0.867}\,/\,\textbf{0.906}\\
\hline
\end{tabular}
  \begin{tablenotes} 
\item $\Delta$ denotes the teacher model. \textbf{bold} denotes the best result
\item $L_h$ indicates $L_{\text{hard}}$; $L_{s}$ indicates $L_{\text{soft}}$; $D_1$ indicates $D_{\text{SaTp}}$; $D_2$ indicates $D_{\text{SpTa}}$; $D_3$ indicates $D_{\text{SaTn}}$; $D_4$ indicates $D_{\text{SnTa}}$.
\end{tablenotes}
  \vspace{-0.2cm}
  \label{tab:KD}
\end{table}

\subsection{Ablation Studies and Insights}
\begin{figure}[t]
  \centering
  \includegraphics[width=\linewidth]{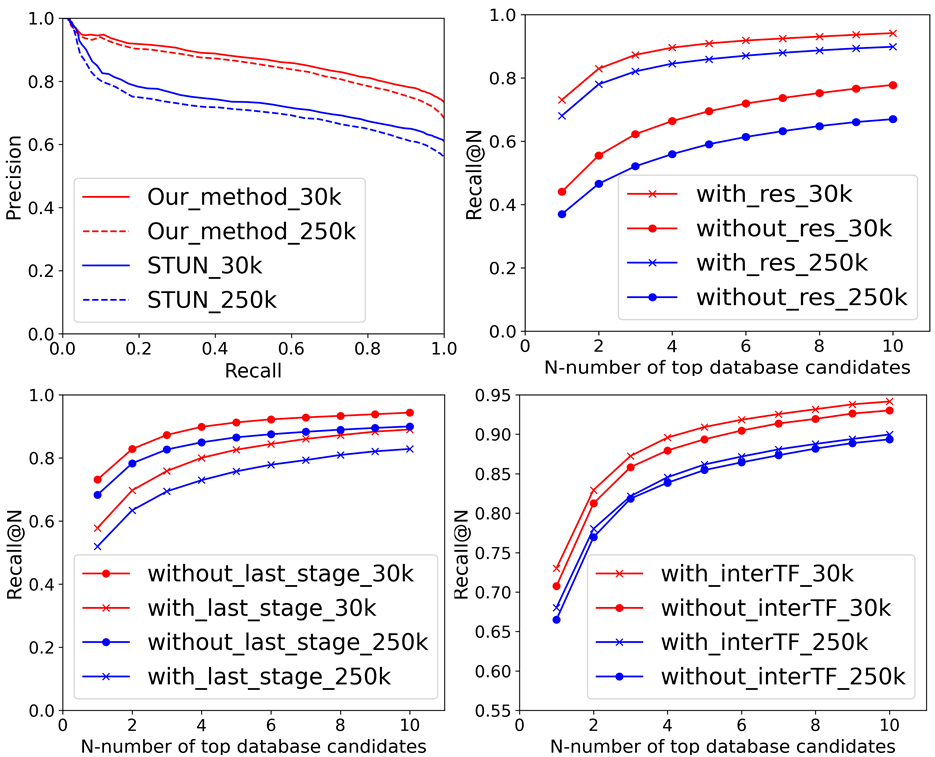}
  \caption{(a) Recall and precision curve for STUN and our method. (b) ablation study on ResNet branch. (c) ablation study on last stage. (d) ablation study on Inter-Transformer Encoder}
  \label{fig:aba}
  \vspace{-0.2cm}
\end{figure}

In this section, we assess the impact of our proposed cross-metric KD and validate our designs on the architecture and our KD strategies. In~\cref{tab:KD}, we denote $D_{\text{SaTp}}$ as the distances between student anchors and teacher positive samples, and so on. As can be seen, our cross-metric knowledge distillation method enhances place recognition within TSCM. Moreover, the performance of our proposed $L_{\text{cm}} = xx$ is notably superior using $D_{\text{SaTp}}$ and $D_{\text{SpTa}}$, compared to scenarios where $D_{\text{SpTa}}$, $D_{\text{SpTa}}$, $D_{\text{SnTa}}$, and $D_{\text{SaTn}}$ are all included, or when either $D_{\text{SpTa}}$, $D_{\text{SaTp}}$, $D_{\text{SnTa}}$ or $D_{\text{SpTa}}$, $D_{\text{SaTp}}$, $D_{\text{SaTn}}$ are added. This substantiates our hypotheses presented in~\cref{fig:CMKD}. 

We conducted ablation experiments to assess the effectiveness of specific architectural components within our teacher network. This included investigating the impact of retaining the ResNet branch, removing the last ResNet stage, and evaluating the contribution of the Inter-Transformer Encoder. The results of these experiments, depicted in~\cref{fig:aba}, revealed the following key findings: i) Retaining the ResNet branch significantly improved model performance; ii) Removing the final ResNet stage did not lead to performance degradation but rather resulted in improvement; iii) The inclusion of the Inter-Transformer Encoder had a positive impact on model performance. In summary, our modified architecture outperforms the original one.

We conducted ablation experiments on our teacher network, examining the effects of specific architectural components. We analyze the impact of retaining the ResNet branch, removing the last ResNet stage, and evaluating the contribution of the Inter-Transformer encoder. As shown in~\cref{fig:aba}, the results reveal the following insights: i) Keeping the ResNet branch improved model performance significantly; ii) Eliminating the final ResNet stage leads to performance improvement; iii) The inclusion of the Inter-Transformer Encoder positively affected model performance. In sum, all of our design choices contribute positively to VPR tasks.

\subsection{Computational Efficiency and Runtime}
This section provides a comparative analysis of the number of parameters and FLOPS in our teacher model, student model, and STUN's student model, as illustrated in~\cref{tab:para}. 
The results show that our student model, when compared to both our teacher model and STUN's student model, has fewer parameters and lower computational requirements, making it a more lightweight option. \cref{tab:para} also presents the runtime outcomes for TSCM, affirming our third claim that TSCM exhibits remarkable computational efficiency, capable of compressing captured images into descriptors about 1.3\,ms and matching in less than 0.6\,ms per query at a database size of 10\,k within the experimental environment. 

\section{Conclusion}
\label{sec:conclusion}
In this paper, we introduce a novel knowledge distillation technique for visual place recognition called cross-metric knowledge distillation. This method empowers the student network to surpass the teacher network's performance when integrated into our teacher-student model for VPR. We conducted extensive experiments using various datasets and compared with existing methods to validate our claims. The results demonstrate that our approach achieves superior recognition accuracy while maintaining a more lightweight model than the baseline method. Furthermore, our proposed cross-metric knowledge distillation method outperforms conventional knowledge distillation techniques. 
Finally, we evaluated the model's processing speed, confirming its ability to compress captured images into descriptors in 1.3\,ms and achieve matching in less than 0.6\,ms per query with a database size of 10\,k, demonstrating real-time performance.


\bibliographystyle{IEEEtran}

\bibliography{glorified, new}

\end{document}

%% file: stachnisslab-latex.tex
\usepackage{graphics}           
\usepackage{times}              
\usepackage{amsmath}            
\usepackage{amssymb}            
\usepackage{graphicx}
\usepackage{algorithm}
\usepackage[noend]{algpseudocode}
\usepackage{booktabs}
\usepackage{color}
\definecolor{instructioncolor}{rgb}{.5,.5,.5}

\usepackage[font=small]{caption}

\def\eqref#1{Eq.~(\ref{#1})}

\makeatletter
\usepackage{xspace}
\DeclareRobustCommand\onedot{\futurelet\@let@token\@onedot}
\def\@onedot{\ifx\@let@token.\else.\null\fi\xspace}

\def\etal{{et al}\onedot}
\makeatother

\def\etalcite#1{\etal~\cite{#1}}

\usepackage{array}
\newcolumntype{L}[1]{>{\raggedright\let\newline\\\arraybackslash\hspace{0pt}}m{#1}}
\newcolumntype{C}[1]{>{\centering\let\newline\\\arraybackslash\hspace{0pt}}m{#1}}
\newcolumntype{R}[1]{>{\raggedleft\let\newline\\\arraybackslash\hspace{0pt}}m{#1}}